\documentclass[conference]{IEEEtran}
\usepackage[utf8]{inputenc}
\usepackage[T1]{fontenc}
\usepackage{cite}
\usepackage{amsmath,amssymb}
\usepackage{graphicx}
\usepackage{textcomp}
\usepackage{xcolor}
\usepackage{booktabs}
\usepackage{tikz}
\usetikzlibrary{positioning,arrows.meta,shapes.geometric,fit,calc}
\usepackage{url}
\usepackage{balance}
\usepackage{microtype}
\usepackage{enumitem}
\usepackage{orcidlink}
\setlist{nosep,leftmargin=*}

\begin{document}

\title{Persistent and Conversational Multi-Method Explainability for Trustworthy Financial AI}

\author{
\IEEEauthorblockN{Georgios Makridis\,\orcidlink{0000-0002-6165-7239}\IEEEauthorrefmark{1},
Georgios Fatouros\,\orcidlink{0000-0001-6843-089X}\IEEEauthorrefmark{1},
John Soldatos\,\orcidlink{0000-0002-6668-3911}\IEEEauthorrefmark{1},
George Katsis\, \IEEEauthorrefmark{1},
Dimosthenis Kyriazis\,\orcidlink{0000-0001-7019-7214}\IEEEauthorrefmark{2}}
\IEEEauthorblockA{ University of Piraeus, Greece\\
\IEEEauthorrefmark{1}ExpertAI-Lux S.à r.l,
\IEEEauthorrefmark{2}Department of Digital Systems\\
\{george.makridis, george.fatouros, john.soldatos, george.katsis\}@expertai-lux.com, dimos@unipi.gr}
}

\maketitle

\begin{abstract}
Financial institutions increasingly require AI explanations that are persistent, cross-validated across methods, and conversationally accessible to human decision-makers.
We present an architecture for human-centered explainable AI in financial sentiment analysis that combines three contributions.
First, we treat XAI artifacts---LIME feature attributions, occlusion-based word importance scores, and saliency heatmaps---as persistent, searchable objects in distributed S3-compatible storage with structured metadata and natural-language summaries, enabling semantic retrieval over explanation history and automatic index reconstruction after system failures.
Second, we enable multi-method explanation triangulation, where a retrieval-augmented generation (RAG) assistant compares and synthesizes results from multiple XAI methods applied to the same prediction, allowing users to assess explanation robustness through natural-language dialogue.
Third, we evaluate the faithfulness of generated explanations using automated checks over grounding completeness, hallucinated claims, and method-attribution behavior.
We demonstrate the architecture on an EXTRA-BRAIN financial sentiment analysis pipeline using FinBERT predictions and present evaluation results showing that constrained prompting reduces hallucination rate by 36\% and increases method-attribution citations by 73\% compared to naive prompting.
We discuss implications for trustworthy, human-centered AI services in regulated financial environments.
\end{abstract}


\section{Introduction}

Explainable AI (XAI) has become a regulatory and practical necessity in financial services, driven by the EU AI Act~\cite{eu_ai_act2024} and growing institutional demand for auditable decision support~\cite{cfa_xai2025}.
Financial sentiment analysis---classifying news headlines as positive, negative, or neutral---is a canonical use case where transformer-based models such as FinBERT~\cite{araci2019finbert} achieve high accuracy but operate as opaque classifiers.
Prior work~\cite{fatouros2023chatgpt_sentiment} demonstrated that LLM-based approaches can substantially outperform FinBERT in financial sentiment classification, yet the resulting predictions remain equally opaque to end-users.
Post-hoc explanation methods including LIME~\cite{ribeiro2016lime}, occlusion-based word importance, and attention analysis are routinely applied to such models~\cite{makridis2024xai_timeseries}, yet three fundamental challenges remain unaddressed in practice.

\textbf{Challenge 1: Explanation ephemerality.}
In current XAI pipelines, explanations are generated on demand, rendered as static visualizations, and discarded.
They are not stored as persistent, searchable records and cannot be queried across sessions, compared over time, or reconstructed after system failures.
This makes compliance auditing and longitudinal explanation analysis impractical---a concern particularly acute in digital finance, where regulatory requirements demand auditable decision trails~\cite{eu_ai_act2024}.
While provenance-aware frameworks~\cite{zhang2024pxai, kale2023provenance} and XAI-as-a-service platforms~\cite{wang2024xaiport, singh2026xaas} have begun treating explanations as managed artifacts, none provide a fully persistent, semantically searchable explanation store with automatic reconstruction.

\textbf{Challenge 2: Single-method fragility.}
\cite{krishna2024disagreement} formally demonstrated that popular XAI methods---LIME, KernelSHAP, Integrated Gradients, attention---routinely disagree on the same instance.
Recent FinBERT-specific evaluation confirms this on financial text, finding that LIME, Integrated Gradients, and attention rollout produce divergent attribution rankings~\cite{finbert_faithfulness2025}.
If a user sees only one explanation method, they receive an incomplete and potentially misleading picture of model behavior.

\textbf{Challenge 3: Conversational XAI faithfulness.}
Conversational XAI systems~\cite{slack2023talktomodel, shen2023convxai, feldhus2023interrolang, wang2024llmcheckup} allow users to query explanations in natural language.
However,~\cite{he2025conversational} demonstrated that LLM-driven explanation chat can amplify over-reliance through an ``illusion of explanatory depth''---the assistant sounds authoritative regardless of whether its claims are grounded in actual XAI evidence.
Without automated faithfulness verification~\cite{es2024ragas, ru2024ragchecker}, fluent but unfaithful explanation narratives erode trust rather than building it.

\textbf{Contributions.}
We present a microservice architecture that addresses all three challenges simultaneously:
\begin{enumerate}
    \item \textbf{Persistent XAI artifact store}: Explanation outputs are stored as structured metadata in S3-compatible object storage with natural-language summaries, enabling semantic vector search over explanation history and automatic index reconstruction after container restarts (Section~\ref{sec:arch}).
    \item \textbf{Multi-method explanation triangulation}: LIME, occlusion, and saliency are applied to the same prediction and their results are jointly indexed, enabling a RAG chatbot to compare, contrast, and synthesize them in response to user questions (Section~\ref{sec:triangulation}).
    \item \textbf{Faithfulness-constrained RAG}: Generated explanations are grounded in retrieved XAI artifacts through constrained prompting and evaluated using automated faithfulness metrics (Section~\ref{sec:faithfulness}).
\end{enumerate}

The system is developed within the EXTRA-BRAIN Horizon Europe project for digital finance and demonstrated on FinBERT-based news sentiment classification with per-sample and dataset-level XAI capabilities.

\section{Related Work}
\label{sec:related}

\textbf{XAI for Financial NLP.}
Post-hoc explanation of financial transformer models has been extensively studied.
In~\cite{rizinski2024xlex} SHAP was used over FinBERT to automatically induce explainable finance lexicons.
In our prior work~\cite{makridis2024xai_timeseries} we extended XAI to time-series classification using image highlight methods (LIME and Grad-CAM) applied to 2D plot representations---a cross-modal approach that informs our saliency-based vision XAI.
In~\cite{fatouros2023chatgpt_sentiment} the authors benchmarked ChatGPT against FinBERT for financial sentiment, showing 35\% improved classification but highlighting the need for interpretability of LLM-based predictions.
\cite{makridis2025virtualxai} proposed VirtualXAI, a framework that integrates quantitative XAI benchmarking with qualitative user assessments through GPT-generated virtual personas---establishing the user-centric evaluation methodology that informs our design.
A comprehensive study applying five XAI methods---LIME, SHAP, attention, Integrated Gradients, and Grad-CAM---in parallel to the same sentiment model~\cite{absa_five_xai2024} provides the closest blueprint for multi-method triangulation.
The 2025 FinBERT faithfulness audit~\cite{finbert_faithfulness2025} quantified deletion-curve faithfulness across LIME, Integrated Gradients, and attention rollout, finding LIME most faithful on financial text.
Our work extends this line by exposing multi-method results interactively rather than as static evaluation tables.

\textbf{Conversational XAI.}
TalkToModel~\cite{slack2023talktomodel} pioneered interactive natural-language explanations, outperforming dashboard XAI in user studies.
ConvXAI~\cite{shen2023convxai} unified heterogeneous explanation types under conversational design principles, while InterroLang~\cite{feldhus2023interrolang} extended the approach to NLP models.
More recent systems such as LLMCheckup~\cite{wang2024llmcheckup} and Explingo~\cite{zytek2024explingo} replaced rule-based intent parsing with LLM orchestration; a comprehensive survey~\cite{beyond_oneshot2025} maps the design space across 15 core systems.
\cite{makridis2025humaine_chatbot} proposed HumAIne-Chatbot, a personalized conversational AI with reinforcement learning that adapts explanation delivery to user profiles.
The authors of~\cite{he2025conversational} found that LLM-driven explanation chat can amplify over-reliance---motivating our faithfulness-constrained design.

\textbf{RAG Faithfulness Evaluation.}
RAGAs~\cite{es2024ragas} introduced reference-free metrics for faithfulness, relevance, and context quality.
ARES~\cite{saad2024ares} fine-tuned lightweight judges with prediction-powered inference.
RAGTruth~\cite{niu2024ragtruth} provided word-level hallucination annotations for supervised detection.
RAGChecker~\cite{ru2024ragchecker} decomposed evaluation into claim-level faithfulness and noise sensitivity metrics.
FActScore~\cite{min2023factscore} established atomic-level groundedness scoring.
We adapt these principles to the XAI domain, where ``faithfulness'' means that the chatbot does not invent features, misattribute explanation methods, or fabricate confidence values.

\textbf{Persistent XAI Artifacts.}
This is the sparsest area and our strongest novelty claim.
\cite{zhang2024pxai} proposed provenance-enabled XAI (PXAI) using lineage graphs as queryable artifacts.
\cite{singh2026xaas} proposed Explainability-as-a-Service with semantic-similarity caching for edge AI.
The principle of treating AI-generated artifacts as persistent, searchable, and personalizable records has been explored in adjacent domains such as education, where~\cite{makridis2024fairylandai} demonstrated persistent personalized content generation with LLMs, establishing the design pattern we extend to XAI artifacts.
Our approach differs from all prior work by combining persistent S3 storage with automatic semantic rehydration of an in-memory vector index, enabling both bulk dataset explanations and per-sample XAI results to survive system failures and be searched conversationally.

\section{System Architecture}
\label{sec:arch}

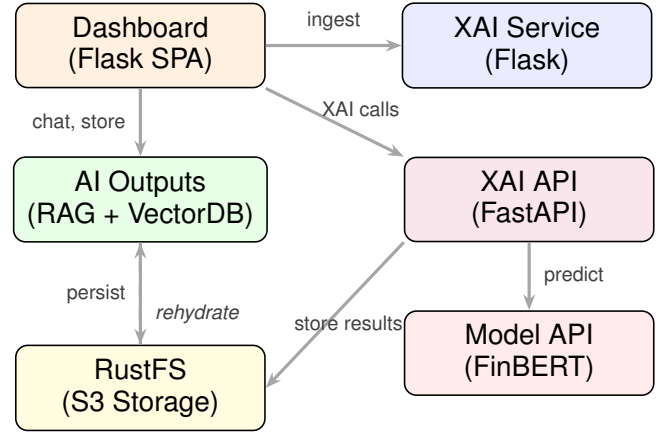
\begin{figure}[t]
\centering
\resizebox{\columnwidth}{!}{
\begin{tikzpicture}[
    node distance=0.6cm and 0.8cm,
    box/.style={draw, rounded corners=3pt, minimum width=2.2cm,
                minimum height=0.65cm, font=\scriptsize\sffamily,
                align=center, fill=#1},
    box/.default=blue!8,
    arrow/.style={-{Stealth[length=4pt]}, thick, color=gray!70},
    label/.style={font=\tiny\sffamily, color=gray!50!black},
]
\node[box=orange!12] (dash)   {Dashboard\\(Flask SPA)};
\node[box, right=1.2cm of dash]  (xaisvc) {XAI Service\\(Flask)};
\node[box=green!10, below=of dash] (aiout)  {AI Outputs\\(RAG + VectorDB)};
\node[box=purple!10, right=1.2cm of aiout] (xaiapi) {XAI API\\(FastAPI)};
\node[box=red!8,    below=of xaiapi]       (model)  {Model API\\(FinBERT)};
\node[box=yellow!15, below=of aiout, yshift=-0.3cm] (rustfs) {RustFS\\(S3 Storage)};

\draw[arrow] (dash) -- node[label, above] {ingest}      (xaisvc);
\draw[arrow] (dash) -- node[label, left]  {chat, store} (aiout);
\draw[arrow] (dash.south east) -- node[label, right, pos=0.3] {XAI calls} (xaiapi.north west);
\draw[arrow] (xaiapi) -- node[label, right] {predict}       (model);
\draw[arrow] (aiout)  -- node[label, left]  {persist}       (rustfs);
\draw[arrow] (xaiapi.south west) -- node[label, below, pos=0.4] {store results} (rustfs.east);
\draw[arrow, dashed] (rustfs) -- node[label, right, pos=0.3] {\textit{rehydrate}} (aiout);
\end{tikzpicture}
}
\caption{System architecture. Solid arrows show runtime data flow; the
dashed arrow represents automatic vector index reconstruction from
persistent metadata on container restart.}
\label{fig:arch}
\end{figure}

The system consists of five Docker microservices orchestrated via Docker
Compose, communicating over a shared network with a remote S3-compatible
object store (RustFS) for persistent storage (Fig.~\ref{fig:arch}).

\textbf{Dashboard} (Flask\,+\,JavaScript SPA) serves the web UI with two
primary tabs: \textit{Dataset Analysis} for bulk model output visualization
across the full dataset, and \textit{Per-Sample XAI} for individual
prediction explanations. It proxies all API calls and handles authentication
via Keycloak OIDC or legacy mode.

\textbf{XAI Service} (Flask) performs bulk data ingestion, preprocessing, and
dataset-level statistical visualization (sentiment distributions, keyword
analysis, asset-level box plots). It holds the ingested data frame in memory
and provides a per-sample browsing interface to the Dashboard.

\textbf{XAI API} (FastAPI) implements the per-sample explanation algorithms:
occlusion-based word importance, LIME~\cite{ribeiro2016lime} text
explanations, and occlusion-based saliency maps for vision-based tasks.
For each request it retrieves the relevant data from the artifact store,
calls the Model API---potentially hundreds of times for perturbation-based
methods---and persists the full result as structured JSON or PNG back to the
store.

\textbf{Model API} serves modular prediction endpoints for the supported
domains: sentiment classification~\cite{araci2019finbert} and, as an
extensibility demonstration, object detection. This component is designed
for straightforward replacement with production models.

\textbf{AI Outputs} (Flask) is the RAG engine. It maintains an in-memory
vector database (SimpleVectorDB with OpenAI embeddings) and conversation
history. It supports question answering over stored explanation artifacts,
indexing of newly generated XAI summaries, and persistence of plot images
with associated metadata.

\subsection{Persistent XAI Artifact Store}
\label{sec:store}

The core architectural innovation is treating every explanation output as a
persistent, searchable artifact rather than an ephemeral visualization.
The artifact store organizes outputs across logically separated buckets for
rendered visualization images, structured metadata records, and uploaded or
exported datasets. Domain-specific results---text-based attribution scores
and vision saliency maps---are stored in dedicated result buckets, keeping
the provenance trail clean across modalities.

Each metadata record follows a consistent schema that captures provenance
fields (model identifier, XAI method, timestamp) alongside storage
references and a natural-language summary object used for semantic retrieval:

{\small
\begin{verbatim}
{"plot_type": "text_occlusion",
 "title":     "Occlusion — sample #5",
 "summary_for_rag": {
   "text":     "Target: positive. Top words:
                outperformer (+0.245), growth (+0.156)",
   "keywords": ["occlusion", "positive"],
   "numeric_facts": {"baseline": 0.912}},
 "provenance": {"model":      "finbert",
                "xai_method": "occlusion"}}
\end{verbatim}
}

\noindent The \texttt{plot\_type} field is set at ingestion time by the
explanation module and acts as a method tag used during retrieval; any new
XAI method can be integrated by registering a new type value without
modifying the storage schema.

\subsection{Automatic Rehydration}
\label{sec:rehydration}

The vector database resides in RAM and is lost on container restart.
A lightweight rehydration mechanism detects an empty collection for the
requesting user on the first conversational query after restart: it scans
all stored metadata records for that user, extracts the natural-language
summary field from each, computes its embedding, and inserts it into the
vector index. This process is transparent to the user---the first query
after a restart incurs a one-time
latency\footnote{Measured on our deployment: 0.8\,s for 15 artifacts,
1.4\,s for 50 artifacts, 2.1\,s for 100 artifacts (single-threaded,
remote RustFS over HTTPS).} after which all subsequent searches operate
from RAM at sub-millisecond latency.

\section{Multi-Method Explanation Triangulation}
\label{sec:triangulation}

A key insight motivating our design is that multiple XAI methods applied to
the same prediction often disagree~\cite{krishna2024disagreement}.
Rather than resolving this disagreement computationally, we expose it to the
user through conversational interaction, enabling what we term
\textit{explanation triangulation}.

For text sentiment classification, the system provides two complementary
per-sample methods:

\textbf{Occlusion.}
Each word is masked individually and the model re-queried; the importance of
word~$w_i$ is the drop in target-class probability:
$\text{imp}(w_i) = P(c^* \mid \mathbf{x}) - P(c^* \mid \mathbf{x}_{\setminus i})$.
This produces a complete attribution over all words, sorted by absolute
importance.

\textbf{LIME.}
A neighbourhood of perturbed texts is generated by randomly removing words,
the model is queried on all perturbations, and a local linear model is
fitted. The coefficients of the top-$k$ features constitute the
explanation~\cite{ribeiro2016lime}.

Both methods are applied to the same sample, and their results are
independently indexed in the vector database with method-specific type tags.
When the user asks the RAG chatbot a comparative question such as
\textit{``Do the XAI methods agree on the most important words?''}, the
retrieval step returns both artifacts (matched by semantic similarity), and
the constrained prompt (Section~\ref{sec:faithfulness}) generates a response
that explicitly cites which method produced which attribution, surfaces
agreements and disagreements, and avoids conflating the two.

For vision tasks, the system provides occlusion-based saliency: a patch
slides across the image, the model is re-queried at each position, and the
confidence drop is mapped to a heat map. This component serves as an
extensibility demonstration---validating that the persistent-artifact and RAG
architecture generalises beyond NLP to vision tasks---and is not covered by
the quantitative evaluation, which focuses on the financial sentiment use
case.

An illustrative example from our financial sentiment dataset: for the
headline \textit{``BioNTech forecasts strong growth in oncology pipeline''},
occlusion identified \textit{strong} (importance $+0.312$) and \textit{growth}
($+0.287$) as the top contributors, while LIME ranked \textit{growth}
($+0.289$) first and \textit{strong} ($+0.195$) third, with
\textit{forecasts} ($+0.201$) intervening.
The chatbot synthesises this as: \textit{``Both methods agree that `growth'
strongly supports the positive prediction. However, occlusion assigns higher
importance to `strong' than LIME does, while LIME elevates `forecasts' as a
contributing factor that occlusion considers less significant.''}
This cross-validation surfaces explanation robustness---or its absence---
through natural dialogue rather than requiring the user to compare static
tables.

\section{Faithfulness Evaluation}
\label{sec:faithfulness}

Following the finding that LLM-driven XAI chat can amplify
over-reliance~\cite{he2025conversational}, we implement a faithfulness
evaluation pipeline that measures whether generated explanations are grounded
in the actual XAI artifacts retrieved by the RAG system.

\subsection{Constrained vs.\ Naive Prompting}

We compare two prompting strategies:

\textbf{Naive baseline:} The retrieved XAI documents are placed in the LLM
context with a generic instruction: \textit{``Answer the user's question
based on the following context.''}

\textbf{Constrained prompt:} The system prompt explicitly requires:
(i)~cite specific XAI artifacts by method and explanation type;
(ii)~distinguish between LIME and occlusion results when both are present;
(iii)~include numeric values (confidence scores, importance weights) directly
from the retrieved documents;
(iv)~state when evidence is insufficient rather than speculating.

\subsection{Faithfulness Metrics}

We adapt RAG faithfulness principles~\cite{es2024ragas, ru2024ragchecker,
min2023factscore} to the XAI domain with three automated rule-based checks
applied to each generated response:

\begin{itemize}
    \item \textbf{Grounding completeness}: fraction of numeric claims in the
    response that can be traced to a retrieved XAI artifact (word importance
    scores, confidence values, or dataset statistics).
    \item \textbf{Hallucination rate}: fraction of feature names mentioned as
    ``important'' in the response that do not appear in any retrieved document.
    \item \textbf{Citation behavior}: number of explicit method attributions
    (e.g., ``according to LIME\ldots'', ``the occlusion analysis
    shows\ldots'') per response.
\end{itemize}

\subsection{Results}

Table~\ref{tab:faithfulness} summarizes results from 30 evaluation queries
spanning single-method questions (10 queries, e.g., ``What were the most
important words?''), comparative multi-method questions (5 queries,
e.g., ``Do the methods agree?''), adversarial probes (5 queries referencing
non-existent features), and dataset-level questions (10 queries about
distributions and statistics).
The ground truth comprises 11 word importance scores (4 top occlusion tokens,
7 top LIME tokens, with 2 overlapping) and 15 distinct explanation types from
the Vector DB.

\begin{table}[t]
\centering
\caption{Faithfulness evaluation: constrained vs.\ naive prompting
         ($n$\,=\,30 queries)}
\label{tab:faithfulness}
\small
\begin{tabular}{lcc}
\toprule
\textbf{Metric} & \textbf{Naive} & \textbf{Constrained} \\
\midrule
Hallucination rate ($\downarrow$) & 0.14 & \textbf{0.09} \\
Citations per response ($\uparrow$) & 5.1 & \textbf{8.9} \\
Grounding completeness ($\uparrow$) & \textbf{0.49} & 0.37$^*$ \\
\bottomrule
\end{tabular}
\vspace{2pt}

{\footnotesize $^*$Constrained prompting produces more specific numeric
claims; with limited ground-truth coverage, additional claims reduce the
grounding ratio despite being contextually appropriate. See discussion
in text.}
\end{table}

Constrained prompting reduces the hallucination rate from 14\% to 9\%
(36\% relative improvement), indicating that explicit grounding instructions
effectively prevent the LLM from inventing feature names.
Citation behavior shows the largest effect: 8.9 explicit method attributions
per response versus 5.1 for naive prompting (73\% increase), confirming that
the constrained prompt encourages method-specific language.

Grounding completeness presents a nuanced result: naive prompting (0.49)
outperforms constrained (0.37).
The constrained prompt instructs the LLM to cite exact numeric values,
producing more numeric claims per response.
However, with only 11 word importance scores in the ground truth and no
standalone numeric values, many additional claims cannot be verified, lowering
the ratio.
This highlights that faithfulness evaluation requires comprehensive
ground-truth coverage; future work should persist richer structured metadata
alongside natural-language summaries.

We note that this evaluation uses rule-based automated checks with a modest
test set; a larger-scale evaluation with LLM-based
judges~\cite{saad2024ares} and human annotation would provide stronger
statistical guarantees.

\section{Discussion}
\label{sec:discussion}

\textbf{Practical implications.}
The persistent artifact store enables a new class of XAI workflows:
longitudinal explanation auditing (``how has the model's reliance on word X
changed over the last month?''), cross-user explanation comparison, and
automated regulatory reporting.
Building on prior risk-aware financial AI work~\cite{fatouros2023deepvar},
which demonstrated probabilistic deep learning for portfolio risk assessment,
this architecture adds the explainability layer that regulators increasingly
demand---ensuring that not only the predictions but also their explanations
are persistent, auditable, and recoverable.

\textbf{Relation to the EU AI Act.}
Article~13 of the EU AI Act~\cite{eu_ai_act2024} requires high-risk AI
systems to provide ``sufficiently transparent'' outputs.
Our architecture operationalizes this through persistent, method-attributed,
searchable explanations with provenance metadata, offering a concrete
technical pathway toward compliance.

\textbf{Limitations.}
Our faithfulness evaluation uses rule-based checks rather than learned
judges~\cite{saad2024ares}; extending to LLM-based evaluation would improve
coverage.
The vector database uses simple cosine-similarity search; production
deployment would benefit from approximate nearest-neighbor indices.
While we demonstrate triangulation between two text XAI methods, the
framework naturally extends to SHAP, Integrated Gradients, and
counterfactuals~\cite{rong2024humancentered}.
Finally, a user study following established HCXAI
protocols~\cite{hoffman2023measures, ehsan2024who_xai} would provide
complementary evidence on user trust and comprehension.

\textbf{Broader applicability.}
While demonstrated on financial sentiment, the architecture is domain-agnostic.
The same persistent-artifact, multi-method, faithfulness-constrained RAG
pattern applies to clinical NLP, manufacturing defect
detection~\cite{makridis2022xai_cyber}, or any setting where explanation
auditability, robustness, and conversational accessibility are required.

\section{Conclusion}

We presented a human-centered XAI architecture for financial AI that
addresses three gaps in current practice: explanation ephemerality,
single-method fragility, and conversational faithfulness.
By treating XAI artifacts as persistent, searchable objects with structured
metadata, enabling multi-method explanation triangulation through a RAG
chatbot, and constraining generated explanations with automated faithfulness
checks, the system provides auditable, robust, and accessible explanations
suitable for regulated financial environments.
Evaluation shows that constrained prompting reduces hallucination by 36\%
and increases method-attribution citations by 73\% compared to naive
prompting, while also revealing that comprehensive ground-truth coverage is
essential for meaningful grounding evaluation.
Future work will extend the faithfulness evaluation with LLM-based judges,
conduct controlled user studies with financial analysts, and enrich the
persistent metadata schema to improve ground-truth coverage for automated
evaluation.

\section*{Acknowledgment}
This work has been supported by the EXTRA-BRAIN project, funded by the European Union's Horizon Europe programme under Grant Agreement No.~101135809.

\balance
\bibliographystyle{IEEEtran}
\bibliography{TI2026_references_v5}

@article{krishna2024disagreement,
  author    = {S. Krishna and T. Han and A. Gu and J. Pombra and S. Jabbari and S. Wu and H. Lakkaraju},
  title     = {The Disagreement Problem in Explainable Machine Learning: A Practitioner's Perspective},
  journal   = {Transactions on Machine Learning Research},
  year      = {2024},
  note      = {arXiv:2202.01602}
}

@article{finbert_faithfulness2025,
  title     = {Fine-Tuning and Explaining {FinBERT} for Sector-Specific Financial News: A Reproducible Workflow},
  journal   = {Electronics},
  year      = {2025},
  volume    = {14},
  number    = {23},
  pages     = {4680},
  publisher = {MDPI}
}

@article{rizinski2024xlex,
  author    = {M. Rizinski and others},
  title     = {Sentiment Analysis in Finance: From Transformers Back to eXplainable Lexicons ({XLex})},
  journal   = {IEEE Access},
  year      = {2024}
}

@article{absa_five_xai2024,
  title     = {Explainable Aspect-Based Sentiment Analysis Using Transformer Models},
  journal   = {Big Data and Cognitive Computing},
  year      = {2024},
  volume    = {8},
  number    = {11},
  pages     = {141}
}

@article{slack2023talktomodel,
  author    = {D. Slack and S. Krishna and H. Lakkaraju and S. Singh},
  title     = {Explaining Machine Learning Models with Interactive Natural Language Conversations Using {TalkToModel}},
  journal   = {Nature Machine Intelligence},
  volume    = {5},
  pages     = {873--883},
  year      = {2023}
}

@inproceedings{feldhus2023interrolang,
  author    = {N. Feldhus and Q. Wang and N. Anikina and A. Chopra and T. Oguz and S. M\"{o}ller},
  title     = {{InterroLang}: Exploring {NLP} Models and Datasets through Dialogue-based Explanations},
  booktitle = {Findings of EMNLP},
  year      = {2023}
}

@inproceedings{shen2023convxai,
  author    = {Z. Shen and Q. Huang and K. Wu and T. Huang},
  title     = {{ConvXAI}: Delivering Heterogeneous {AI} Explanations via Conversations to Support Human-{AI} Scientific Writing},
  booktitle = {CSCW Companion},
  year      = {2023}
}

@inproceedings{wang2024llmcheckup,
  author    = {Q. Wang and N. Anikina and N. Feldhus and J. van Genabith and L. Hennig and S. M\"{o}ller},
  title     = {{LLMCheckup}: Conversational Examination of Large Language Models via Interpretability Tools and Self-Explanations},
  booktitle = {NAACL HCINLP Workshop},
  year      = {2024}
}

@inproceedings{zytek2024explingo,
  author    = {A. Zytek and others},
  title     = {Explingo: Explaining {AI} Predictions using Large Language Models},
  booktitle = {IEEE BigData},
  year      = {2024},
  note      = {arXiv:2412.05145}
}

@inproceedings{he2025conversational,
  author    = {G. He and A. Aishwarya and U. Gadiraju},
  title     = {Is Conversational {XAI} All You Need? Human-{AI} Decision Making With a Conversational {XAI} Assistant},
  booktitle = {Proc. ACM IUI},
  year      = {2025}
}

@article{beyond_oneshot2025,
  title     = {Beyond One-Shot Explanations: A Systematic Literature Review of Dialogue-Based {xAI} Approaches},
  journal   = {Artificial Intelligence Review},
  volume    = {58},
  year      = {2025},
  publisher = {Springer}
}

@inproceedings{es2024ragas,
  author    = {S. Es and others},
  title     = {{RAGAs}: Automated Evaluation of Retrieval Augmented Generation},
  booktitle = {EACL Demos},
  year      = {2024}
}

@inproceedings{saad2024ares,
  author    = {J. Saad-Falcon and others},
  title     = {{ARES}: An Automated Evaluation Framework for Retrieval-Augmented Generation Systems},
  booktitle = {NAACL},
  year      = {2024}
}

@inproceedings{niu2024ragtruth,
  author    = {C. Niu and H. Wu and others},
  title     = {{RAGTruth}: A Hallucination Corpus for Developing Trustworthy Retrieval-Augmented Language Models},
  booktitle = {ACL},
  year      = {2024}
}

@inproceedings{ru2024ragchecker,
  author    = {D. Ru and others},
  title     = {{RAGChecker}: A Fine-grained Framework for Diagnosing Retrieval-Augmented Generation},
  booktitle = {NeurIPS Datasets and Benchmarks},
  year      = {2024}
}

@inproceedings{min2023factscore,
  author    = {S. Min and others},
  title     = {{FActScore}: Fine-grained Atomic Evaluation of Factual Precision in Long Form Text Generation},
  booktitle = {EMNLP},
  year      = {2023}
}

@inproceedings{zhang2024pxai,
  author    = {S. Zhang and J. Zhou and B. Ujcich},
  title     = {Provenance-Enabled Explainable {AI}},
  booktitle = {SIGMOD},
  journal   = {PACMMOD},
  volume    = {2},
  number    = {6},
  year      = {2024}
}

@article{kale2023provenance,
  author    = {G. Kale and D. Nguyen and R. Harris and C. Li and S. Zhang and Y. Ma},
  title     = {Provenance Documentation to Enable Explainable and Trustworthy {AI}: A Literature Review},
  journal   = {Data Intelligence},
  volume    = {5},
  number    = {1},
  pages     = {139--162},
  year      = {2023}
}

@inproceedings{wang2024xaiport,
  author    = {Y. Wang and others},
  title     = {{XAIport}: A Service Framework for the Early Adoption of {XAI} in {AI} Model Development},
  booktitle = {ICSE-NIER},
  year      = {2024}
}

@article{singh2026xaas,
  author    = {R. Singh and S. Roy},
  title     = {Scalable Explainability-as-a-Service ({XaaS}) for Edge {AI} Systems},
  journal   = {arXiv:2602.04120},
  year      = {2026}
}

@article{hoffman2023measures,
  author    = {R. R. Hoffman and S. T. Mueller and G. Klein and J. Litman},
  title     = {Measures for Explainable {AI}: Explanation Goodness, User Satisfaction, Mental Models, Curiosity, Trust, and Human-{AI} Performance},
  journal   = {Frontiers in Computer Science},
  volume    = {5},
  year      = {2023}
}

@article{ehsan2024who_xai,
  author    = {U. Ehsan and K. Passi and Q. V. Liao and L. Chan and I. Lee and M. Muller and M. O. Riedl},
  title     = {The Who in {XAI}: How {AI} Background Shapes Perceptions of {AI} Explanations},
  booktitle = {CHI},
  year      = {2024}
}

@article{rong2024humancentered,
  author    = {Y. Rong and T. Leemann and others},
  title     = {Towards Human-Centered Explainable {AI}: A Survey of User Studies for Model Explanations},
  journal   = {IEEE TPAMI},
  year      = {2024}
}

@article{cfa_xai2025,
  author    = {A. Wilson},
  title     = {Explainable {AI} in Finance: Addressing the Needs of Diverse Stakeholders},
  institution = {CFA Institute Research and Policy Center},
  year      = {2025}
}

@article{ribeiro2016lime,
  author    = {M. T. Ribeiro and S. Singh and C. Guestrin},
  title     = {Why Should {I} Trust You?: Explaining the Predictions of Any Classifier},
  booktitle = {KDD},
  year      = {2016}
}

@article{eu_ai_act2024,
  title     = {Regulation ({EU}) 2024/1689 Laying Down Harmonised Rules on Artificial Intelligence ({AI} Act)},
  journal   = {Official Journal of the European Union},
  year      = {2024}
}

@inproceedings{makridis2024xai_timeseries,
  author    = {G. Makridis and G. Fatouros and V. Koukos and D. Kotios and D. Kyriazis and J. Soldatos},
  title     = {{XAI} for Time-Series Classification Leveraging Image Highlight Methods},
  booktitle = {Management of Digital EcoSystems (MEDES 2023)},
  series    = {CCIS},
  volume    = {2022},
  publisher = {Springer},
  year      = {2024},
  doi       = {10.1007/978-3-031-51643-6_28}
}

@article{fatouros2023deepvar,
  author    = {G. Fatouros and G. Makridis and D. Kotios and J. Soldatos and M. Filippakis and D. Kyriazis},
  title     = {{DeepVaR}: A Framework for Portfolio Risk Assessment Leveraging Probabilistic Deep Neural Networks},
  journal   = {Digital Finance},
  volume    = {5},
  pages     = {29--56},
  year      = {2023},
  doi       = {10.1007/s42521-022-00050-0}
}

@article{fatouros2023chatgpt_sentiment,
  author    = {G. Fatouros and J. Soldatos and K. Kouroumali and G. Makridis and D. Kyriazis},
  title     = {Transforming Sentiment Analysis in the Financial Domain with {ChatGPT}},
  journal   = {Machine Learning with Applications},
  volume    = {14},
  pages     = {100508},
  year      = {2023},
  doi       = {10.1016/j.mlwa.2023.100508}
}

@inproceedings{makridis2022xai_cyber,
  author    = {G. Makridis and others},
  title     = {{XAI} Enhancing Cyber Defence Against Adversarial Attacks in Industrial Applications},
  booktitle = {IEEE 5th Intl. Conf. Image Processing Applications and Systems (IPAS)},
  pages     = {1--8},
  year      = {2022}
}

@inproceedings{makridis2025virtualxai,
  author    = {G. Makridis and V. Koukos and G. Fatouros and M. M. Separdani and D. Kyriazis and J. Soldatos},
  title     = {{VirtualXAI}: A User-Centric Framework for Explainability Assessment Leveraging {GPT}-Generated Personas},
  booktitle = {Proc. 21st Intl. Conf. Distributed Computing in Smart Systems and the Internet of Things (DCOSS-IoT)},
  publisher = {IEEE},
  year      = {2025}
}

@article{makridis2025humaine_chatbot,
  author    = {G. Makridis and V. Fragiadakis and H. Oliveira and P. Saraiva and P. Mavrepis and G. Fatouros and D. Kyriazis},
  title     = {{HumAIne-Chatbot}: Real-Time Personalized Conversational {AI} via Reinforcement Learning},
  journal   = {arXiv:2509.04303},
  year      = {2025}
}

@article{makridis2024fairylandai,
  author    = {G. Makridis and A. Oikonomou and V. Koukos},
  title     = {{FairyLandAI}: Personalized Fairy Tales Utilizing {ChatGPT} and {DALL-E} 3},
  journal   = {arXiv:2407.09467},
  year      = {2024}
}

@article{araci2019finbert,
  title={Finbert: Financial sentiment analysis with pre-trained language models},
  author={Araci, Dogu},
  journal={arXiv preprint arXiv:1908.10063},
  year={2019}
}

\end{document}